\documentclass[10pt,twocolumn,letterpaper]{article}

\usepackage[pagenumbers]{cvpr} % To force page numbers, e.g. for an arXiv version
\usepackage{balance}
\usepackage{dsfont}

\usepackage{multibib}
\usepackage{etoolbox}
\makeatletter
\providecommand{\sf@counterlist}{}
\pretocmd{\multicols}{\let\sf@counterlist\@empty}{}{}
\makeatother
\newcites{app}{References}
\usepackage{graphicx}
\usepackage{svg}
\usepackage{booktabs}
\usepackage{rotating}
\usepackage{multirow} 
\usepackage{pifont}
\usepackage{textcomp}
\usepackage{stfloats}
\usepackage{url}
\usepackage{verbatim}
\usepackage{amsmath,amsfonts}
\usepackage{algorithmic}
\usepackage{array}
\usepackage{amssymb}
\usepackage{microtype}
\usepackage{tabularx}
\usepackage[caption=false,font=normalsize,labelfont=sf,textfont=sf]{subfig}

\makeatletter

\newcommand{\Rmnum}[1]{\expandafter\@slowromancap\romannumeral #1@}
\makeatother

\definecolor{cvprblue}{rgb}{0.21,0.49,0.74}
\usepackage[pagebackref,breaklinks,colorlinks,allcolors=cvprblue]{hyperref}

\def\paperID{9661} % *** Enter the Paper ID here
\def\confName{CVPR}
\def\confYear{2025}

\title{RoBiS: \underline{Ro}bust \underline{Bi}nary \underline{S}egmentation for High-Resolution Industrial Images}

\author{
Xurui Li\textsuperscript{1} \quad
Zhongsheng Jiang\textsuperscript{1} \quad
Tingxuan Ai\textsuperscript{1} \quad
Yu Zhou\textsuperscript{1,2,3,$\dagger$} \\
\{xrli\_plus, zsjiang, tingxuanai, yuzhou\}@hust.edu.cn \\
$^1$School of Electronic Information and Communications,
Huazhong University of Science and Technology \\
$^2$ Hubei Key Laboratory of Smart Internet Technology,
Huazhong University of Science and Technology \\
$^3$Artificial Intelligence Research Institute, Wuhan JingCe Electronic Group Co., LTD \\
\textbf{Track \Rmnum{1}---Adapt \& Detect: Robust Anomaly Detection in Real-World Applications}
}

\begin{document}

\maketitle
\begin{abstract}
Robust unsupervised anomaly detection (AD) in real-world scenarios is an important task. Current methods exhibit severe performance degradation on the MVTec AD 2 benchmark due to its complex real-world challenges. To solve this problem, we propose a robust framework RoBiS, which consists of three core modules:
(1) Swin-Cropping, a high-resolution image pre-processing strategy to preserve the information of small anomalies through overlapping window cropping.
(2) The data augmentation of noise addition and lighting simulation is carried out on the training data to improve the robustness of AD model.
We use INP-Former as our baseline, which could generate better results on the various sub-images.
(3) The traditional statistical-based binarization strategy (mean+3std) is combined with our previous work, MEBin (published in CVPR2025), for joint adaptive binarization.
Then, SAM is further employed to refine the segmentation results.
Compared with some methods reported by the MVTec AD 2, our RoBiS achieves a \textbf{29.2\%} SegF1 improvement (from 21.8\% to 51.00\%) on $\textit{TEST}_\text{priv}$ and \textbf{29.82\%} SegF1 gains (from 16.7\% to 46.52\%) on $\textit{TEST}_\text{priv,mix}$.
Code is available at \href{https://github.com/xrli-U/RoBiS}{github link}.
\end{abstract}

\renewcommand{\thefootnote}
\footnotetext{$\dagger$ Corresponding Author.}
\renewcommand{\thefootnote}{\arabic{footnote}}

\section{Introduction}
Unsupervised anomaly detection addresses the critical challenge of identifying deviations from normal patterns using only defect-free training data,
with applications spanning medical diagnostics \cite{TMI2023encoderdecoder}, industrial inspection \cite{arxiv2024visualanomaly, TCYB2024imiad}, and video surveillance \cite{ESWA2018abnormal}.
While current methods achieve near-perfect performance (AUROC~\textgreater~99\%) on benchmark datasets like MVTec AD \cite{cvpr2019mvtec}, their real-world deployment faces fundamental limitations.
Real-world scenarios introduce complex and diverse interferences, such as various lighting conditions, viewpoint changes, and sensor noise.
These challenges expose the fragility of existing approaches that lack explicit mechanisms for handling real-world condition variations, particularly in preserving detection accuracy under significant domain shifts.

The MVTec AD 2 \cite{arxiv2025mvtec2} dataset introduces eight industrial inspection scenarios with 8,004 high-resolution images, establishing a rigorous benchmark for unsupervised anomaly detection.
While training data maintains regular lighting conditions, the test suite comprises three distinct challenges:
(1) Public testset ($\textit{TEST}_\text{pub}$) provides limited annotated samples under mixed illumination for preliminary validation,
(2) Private testset ($\textit{TEST}_\text{priv}$) preserves training-domain conditions for conventional evaluation,
and (3) Mixed private testset ($\textit{TEST}_\text{priv,mix}$) combines seen and unseen lighting conditions to simulate real-world uncertainties.
Current methods achieving more than 99\% AUROC on controlled benchmarks suffer large performance degradation under these complex conditions, exposing the urgent need for robust anomaly detection frameworks.

\textbf{Challenge 1: Robustness on varying scenarios.}
Eight industrial products of the MVTec AD 2 dataset emphasize distinct technical challenges.
Spatial randomness challenges emerge in categories like wall plugs and walnuts, where overlapping objects create complex occlusion patterns.
Material diversity challenges occur in fabric, sheet metal and rice categories, which require texture consistency modeling.
Optical complexities involve reflective surfaces and translucent materials, such as cans, vials and fruit jellies, that distort visual signatures.
Illumination challenges include extreme lighting conditions---backlighting present in vials/fruit jelly, as well as dark-field lighting in metal sheets.
Additionally, grayscale images, high-resolution inputs and non-uniform aspect ratios collectively simulate real-world industrial inspection conditions.

\textbf{Challenge 2: Adaptive binarization.}
Current anomaly detection methods could only generate continuous anomaly maps, where the pixel value indicates the anomaly likelihood.
To conduct an accurate binary discrimination for each pixel, it is necessary to study the adaptive binarization method.
Some researches \cite{iccv2023diffumask,eccv2024dual} in the natural scenarios struggle to be used in the industrial scenarios.
Otsu \cite{TSMC1979otsu} often brings much over-detections due to the normal regions of products also have higher scores than the background.
Using the average of all per-pixel anomaly scores plus three times their standard deviation as the threshold is effective, but it easily overlooks subtle anomalies with low scores.

The above challenges test the robustness and adaptability of unsupervised anomaly detection models in real-world detection scenarios.
Therefore, we design the \textbf{Ro}bust \textbf{Bi}nary \textbf{S}egmentation (\textbf{RoBiS}) to handle these challenges and output precise binarization anomaly segmentation results.

\begin{figure*}[t]
\vspace{-.1in}
\begin{center}
\includegraphics[width=1\textwidth]{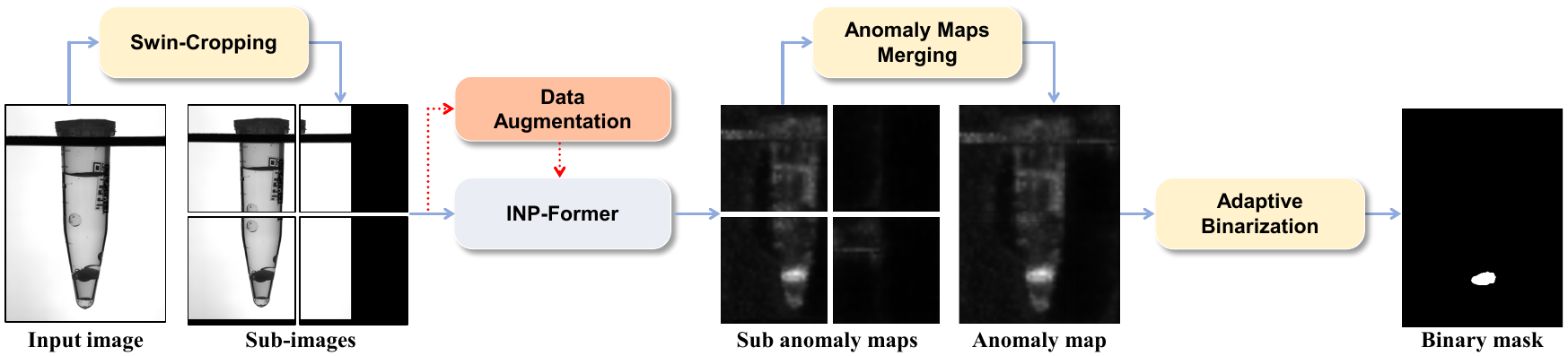}
\vspace{-5mm} 
\caption{
\textbf{The pipeline of our RoBiS.}
This framework contains three important parts:
(1) Swin-Cropping for dataset pre-processing (Sec. \ref{data_preprocess}).
(2) INP-Former \cite{cvpr2025inp-former} to detect anomalies (Sec. \ref{detect_model}).
(3) Anomaly Maps Merging to obtain the anomaly map of original image, and Adaptive Binarization to generate its corresponding binary mask (Sec. \ref{data_postprocess}).
}
\label{fig:pipeline}
\end{center}
\vspace{-10pt}
\end{figure*}

\section{Methodology}
The pipeline of our method is illustrated in Fig. \ref{fig:pipeline}, which consists of three important parts: dataset pre-processing, model design and results post-processing.

\subsection{Dataset Pre-processing}
\label{data_preprocess}
Through the statistical analysis of the MVTec AD 2 \cite{arxiv2025mvtec2} dataset, we observe that all product images maintain high resolutions (ranging from 1400×1900 to 2448×2048).
However, over 20\% of anomalies occupy less than 281 pixels, with the minimal anomaly containing only 5 pixels.
To address this challenge of small anomaly localization, we propose the sliding window-based image cropping strategy, named Swin-Cropping.
This pre-processing strategy systematically crops high-resolution inputs into smaller sub-images for subsequent model training and inference.

The Swin-Cropping strategy employs a 1024×1024 window size to balance computational efficiency and detection accuracy, as excessive partitioning would lead to exponential growth in training time.
To prevent boundary artifacts \cite{icip2008boundary-artifacts} where one anomaly might be divided by adjacent windows, we implement a 10\% overlap between neighboring sub-images.
This design ensures the integrity of large or elongated anomaly patterns during the segmentation process.
To process peripheral regions, zero-padding is applied to residual boundary areas,
maintaining consistent input dimensions for subsequent network operations while preserving spatial relationships in marginal zones.

\subsection{Model Design}
\label{detect_model}
\subsubsection{Approach}
While texture categories (e.g., fabric, rice, and sheet metal) exhibit consistent sub-image patterns, significant inter-subimage variations exist in other product categories.
Traditional one-class anomaly detection (AD) methods \cite{eccv2024glad, eccv2024glass, eccv2024transfusion, cvpr2024realnet} focus on training individual model for each category,
while each category in the MVTec AD 2 dataset contains multiple distinct sub-images.
This architectural constraint prevents effective learning across all sub-image variations.
In contrast, multi-class approaches \cite{cvpr2025inp-former, cvpr2025dinomaly, cvpr2025DeCo-Diff, NIPS2024hemambaad} demonstrate superior learning capacity, particularly in categories like vial and fruit jelly, which have more sub-image discrepancies.
Based on the above analysis, we adopt INP-Former \cite{cvpr2025inp-former} as our baseline, which is the current state-of-the-art multi-class AD framework.
Notice that our method strictly maintains the unsupervised setting and trains one model for each product category.

\subsubsection{Architecture}
Our implementation preserves the original INP-Former architecture while employing a ViT-B-14 backbone initialized with DINOv2-R \cite{iclr2024dino-r} pre-trained weights as the encoder.
To maintain the details in high-resolution, all sub-images from both training and test sets are resized to 518×518 resolution.
In addition, since some anomalies may appear around image boundaries, we discard the center cropping operation to generate more reliable segmentation results.

\subsubsection{Training}
\label{data_aug}
While training images are captured under regular lighting conditions, test scenarios contain significant illumination variations including underexposure, overexposure, and additional light sources.
To enhance model robustness against these illumination variations, we implement an augmentation pipeline applied to 50\% of training samples.
For each sub-image $I_i$, we first inject Gaussian noise with zero mean and standard deviation $\sigma=15$, producing $\overline{I}_i$.
Subsequently, we simulate illumination variations through exposure adjustment to generate the augmented image $\hat{I}_i$ as,
\begin{equation}
\hat{I}_i= \text{clamp}(\overline{I}_i * 2^{\lambda})
\end{equation}
where $\text{clamp}(\cdot)$ constrains pixel values to [0, 255].
And $\lambda \sim \mathcal{U}(-0.2, 0.2)$ controls exposure levels, where positive values induce overexposure while negative values create underexposure effects.
This dual-stage augmentation generates photometrically perturbed samples $\hat{I}_i$ that better approximate real-world testing conditions.

Our training configuration employs the StableAdamW \cite{nips2023stableadamw} optimizer with an initial learning rate of $1\times10^{-3}$ for 200 epochs, coupled with a WarmCosineScheduler for dynamic learning rate adjustment.
The scheduler incorporates 100 warmup iterations to stabilize early training phases, gradually decaying the learning rate to a final value of $1\times10^{-4}$ through cosine annealing.

\begin{table*}[!t]
    \vspace{-.4in}
  \centering
      \setlength\tabcolsep{3pt}
        \resizebox{1.0\linewidth}{!}{
    \begin{tabular}{llcccccccc}
    \toprule
    & Object & PatchCore \cite{CVPR2022patchcore} & RD \cite{cvpr2022rd} & RD++ \cite{cvpr2023rd++} & EfficientAD \cite{wacv2024efficientad} & MSFlow \cite{TNNLS2024msflow} & SimpleNet \cite{cvpr2023simplenet} & DSR \cite{DSR} & RoBiS (Ours) \\
    \midrule 
    & Can & 0.3~/~0.1 & 0.1~/~0.1 & 0.1~/~0.1 & 0.8~/~0.1 & \textbf{5.0}~/~0.1 & 0.6~/~0.1 & 0.4~/~0.1 & \underline{1.86}~/~\textbf{0.84} \\
    & Fabric & 11.5~/~9.8 & 2.6~/~2.2 & 2.9~/~2.3 & 7.6~/~1.0 & \underline{22.0}~/~4.1 & 21.6~/~\underline{10.2} & 7.9~/~5.0 & \textbf{87.46}~/~\textbf{73.37} \\
    & Fruit Jelly & 8.7~/~8.2 & 22.5~/~22.7 & 26.9~/~26.7 & 20.8~/~18.2 & \underline{47.6}~/~\underline{38.1} & 25.1~/~23.0 & 17.9~/~17.2 & \textbf{53.63}~/~\textbf{52.62} \\
    & Rice & 3.8~/~\underline{4.2} & 7.0~/~3.9 & 9.5~/~2.9 & 15.0~/~0.5 & \underline{19.1}~/~1.8 & 11.6~/~1.0 & 1.5~/~1.4 & \textbf{63.86}~/~\textbf{63.23} \\
    & Sheet Metal & 1.8~/~1.1 & \underline{41.3}~/~\underline{39.2} & 40.9~/~37.7 & 9.3~/~3.8 & 13.0~/~7.6 & 14.6~/~2.8 & 13.9~/~14.4 & \textbf{70.98}~/~\textbf{70.92} \\
    & Vial & 2.3~/~2.2 & 28.0~/~\underline{28.3} & 28.2~/~22.8 & 30.5~/~26.5 & 23.3~/~6.2 & \underline{31.9}~/~17.5 & 28.2~/~27.9 & \textbf{48.73}~/~\textbf{48.83} \\
    & Wall Plugs & 0.0~/~0.0 & 1.9~/~0.8 & 1.3~/~\underline{0.9} & \underline{4.4}~/~0.3 & 0.1~/~0.2 & 1.0~/~0.3 & 0.4~/~0.4 & \textbf{14.38}~/~\textbf{3.40} \\
    & Walnuts & 1.2~/~1.3 & 41.2~/~36.7 & 44.1~/~\underline{40.5} & 34.6~/~13.3 & \underline{44.5}~/~14.3 & 35.2~/~14.3 & 17.0~/~9.6 & \textbf{67.13}~/~\textbf{58.94} \\
    \midrule 
    & Mean & 3.7~/~3.4 & 18.1~/~\underline{16.7} & 19.2~/~\underline{16.7} & 15.4~/~8.0 & \underline{21.8}~/~9.0 & 17.7~/~8.7 & 10.9~/~9.5 & \textbf{51.00}~/~\textbf{46.52} \\
    \bottomrule
    \end{tabular}
    }
  \vspace{-0.5em}
  \caption{Segmentation $F_1$ score (in \%) on binarized images for $\textit{TEST}_\text{priv}$/$\textit{TEST}_\text{priv,mix}$ set on the MVTec AD 2 dataset. The best-performing result is in bold, the second-best result is underlined.}
  \label{tab:main_comp}
\end{table*}

\subsection{Results Post-processing}
\label{data_postprocess}
During inference, our method generates anomaly maps for each sub-image, where each pixel value indicates anomaly likelihood.
To obtain final binary segmentation masks for original high-resolution images, we implement a two-stage post-processing workflow.

\ding{182} \textbf{Anomaly Maps Merging.}
The spatial merging of anomaly maps begins with precise coordinate alignment using positional metadata recorded during Swin-Cropping preprocessing.
Each sub-image's anomaly map is repositioned in the original high-resolution coordinate system through bilinear interpolation.
For overlapping regions between adjacent sub-images (10\% overlap ratio), we compute pixel-wise probabilistic averages.

\ding{183} \textbf{Adaptive Binarization.}
For anomaly map $A_i$, the traditional binarization uses the average of all per-pixel anomaly scores plus three times their standard deviation as the threshold to produce the binary mask $\overline{M}_i$.
However, this approach may lead to missing detection for subtle anomalies, which have lower anomaly scores.
To mitigate this, we employ our MEBin proposed in AnomalyNCD \cite{cvpr2025anomalyncd} (CVPR 2025), which adaptively determines optimal thresholds through stable connected-component analysis, generating the mask $\hat{M}_i$.
Then we calculate the coarse segmentation mask through the logical OR operation as,
\begin{equation}
M_i=\overline{M}_i \cup \hat{M}_i
\end{equation}

To enhance segmentation precision, we develop a SAM-based refinement module (SAM-Finer) to process the coarse segmentation mask $M_i$.
First, we extract the minimum bounding box for each abnormal region of $M_i$.
Then each bounding box is considered as a prompt of Segment Anything (SAM) \cite{cvpr2023sam} to guide finer segmentation.
Since the SAM decoder generates three confidence-ranked masks for each prompt, we combined them through logical OR operations to minimize false negatives for the fabric and walnuts categories.
We only use the mask with the largest confidence for the other categories.
Final binary mask $\mathbf{M}_i$ is obtained through the spatial composition of all refined abnormal regions.
The above SAM-Finer is used on all categories except rice.
In addition, we use SAM-Huge architecture for $\textit{TEST}_\text{priv}$ and SAM-Base for $\textit{TEST}_\text{priv,mix}$.

\subsection{Dataset \& Evaluation}
We conduct experiments on the industrial dataset MVTec AD 2 \cite{arxiv2025mvtec2}.
The complete dataset pre-processing workflow and technical details of our data augmentation strategies are documented in Sec. \ref{data_preprocess} and Sec. \ref{data_aug}, respectively.

We strictly adhere to the official challenge's evaluation paradigm, employing pixel-level F1-score (SegF1) as the primary metric for assessing anomaly detection performance.
Conventional AD methods only produce continuous anomaly maps.
To calculate the SegF1 metric, binarizing these anomaly maps is necessary.

\section{Results}

\subsection{Quantitative Results}
As reported in Table \ref{tab:main_comp}, our method achieves superior segmentation performance on MVTec AD 2's challenging test sets,
attaining SegF1 scores of 51.00\% and 46.52\% on $\textit{TEST}_\text{priv}$ and $\textit{TEST}_\text{priv,mix}$ respectively.
This represents significant improvements of \textbf{29.2\%} and \textbf{29.82\%} over state-of-the-art competitors including PatchCore \cite{CVPR2022patchcore}, RD \cite{cvpr2022rd}, RD++ \cite{cvpr2023rd++}, EfficientAD \cite{wacv2024efficientad}, MSFlow \cite{TNNLS2024msflow}, SimpleNet \cite{cvpr2023simplenet} and DSR \cite{DSR}, as officially benchmarked in MVTec AD 2.
Our comprehensive evaluation further reveals consistent advantages across multiple metrics (Table \ref{tab:detail_results}), such as AucPro\_0.05 and ClassF1.
Notably, while the challenge ranking prioritizes SegF1, our method's balanced performance across all metrics (ClassF1: 79.62\%/79.86\%, AucPro\_0.05: 67.25\%/59.65\%) confirms its robustness and adaptability for different industrial scenes.
These metrics are all calculated by the official platform\footnote{\href{https://benchmark.mvtec.com/vand-leaderboard}{https://benchmark.mvtec.com/vand-leaderboard}} of the challenge.

\begin{table}[!t]
    \centering
  \label{tab:our_results}
  \setlength{\tabcolsep}{0.8mm}{
    \begin{tabular}{lccc}
    \toprule
    \multirow{2}{*}{Object} & AucPro\_0.05 & ClassF1 & SegF1 \\
    ~ & (private) & (private) & (private) \\
    \midrule
    Can & 30.28 & 60.93 & 1.86 \\
    Fabric & 79.45 & 83.79 & 87.46 \\
    Fruit Jelly & 74.46 & 87.35 & 53.63 \\
    Rice & 62.27 & 72.00 & 63.86 \\
    Sheet Metal & 75.51 & 87.68 & 70.98 \\
    Vial & 76.81 & 84.61 & 48.73 \\
    Wall Plugs & 62.20 & 75.20 & 14.38 \\
    Walnuts & 77.05 & 85.42 & 67.13 \\
    \midrule[0.2pt]
    Mean & 67.25 & 79.62 & 51.00 \\
    \midrule[0.8pt]
    \multirow{2}{*}{Object} & AucPro\_0.05 & ClassF1 & SegF1 \\
    ~ & (priv\_mixed) & (priv\_mixed) & (priv\_mixed) \\
    \midrule
    Can & 20.03 & 65.04 & 0.84 \\
    Fabric & 79.27 & 83.80 & 73.37 \\
    Fruit Jelly & 74.11 & 87.55 & 52.62 \\
    Rice & 63.89 & 73.45 & 63.23 \\
    Sheet Metal & 73.54 & 86.69 & 70.92 \\
    Vial & 69.59 & 85.77 & 48.83 \\
    Wall Plugs & 24.77 & 72.66 & 3.40 \\
    Walnuts & 72.00 & 83.95 & 58.94 \\
    \midrule[0.2pt]
    Mean & 59.65 & 79.86 & 46.52 \\
    \bottomrule
    \end{tabular}
    }
\caption{More quantitative results on $\textit{TEST}_\text{priv}$/$\textit{TEST}_\text{priv,mix}$ set.}
\label{tab:detail_results}
\end{table}

\begin{figure*}[t]
\vspace{-.1in}
\begin{center}
\includegraphics[width=0.96\textwidth]{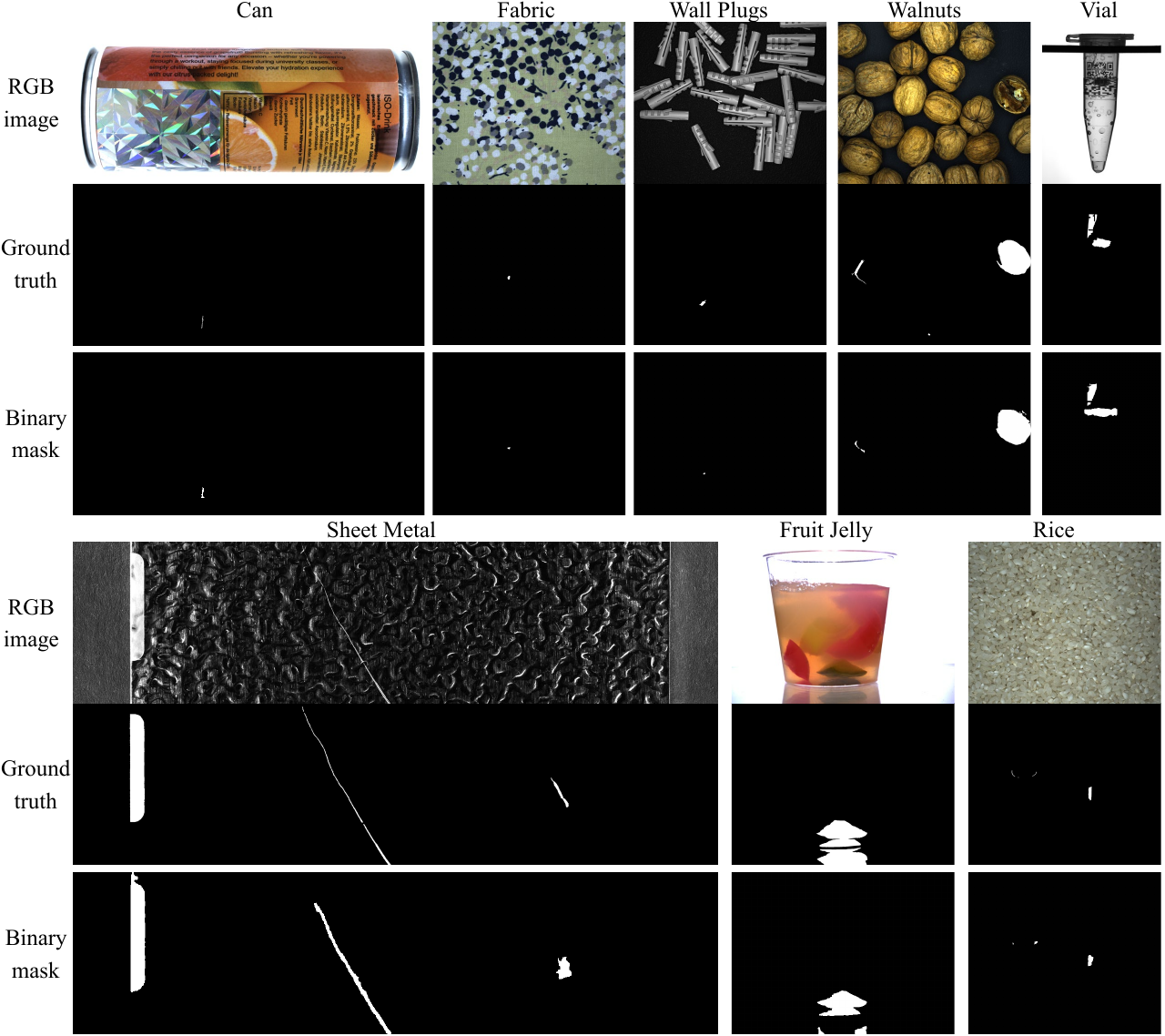}
\vspace{-1mm} 
\caption{
Visualization of anomaly segmentation results on the $\textit{TEST}_\text{pub}$ set of MVTec AD 2 dataset.
}
\label{fig:vis}
\end{center}
\vspace{-10pt}
\end{figure*}

\subsection{Qualitative Results}
Since the ground truth masks of $\textit{TEST}_\text{priv}$ and $\textit{TEST}_\text{priv,mix}$ sets are unavailable,
we visualize the binary mask of $\textit{TEST}_\text{pub}$ set output by our method in Fig. \ref{fig:vis}.
Our method demonstrates effective anomaly localization across most of the MVTec AD 2 product categories.
While our method successfully identifies both structural and logical anomalies, architectural constraints inherent to anomaly detection models, particularly the need for successive downsampling to achieve sufficient receptive fields.
This manifests in binary masks that are larger than ground truth annotations.
Although SAM integration provides partial refinement, the 64×64 feature resolution limit of SAM's encoder prevents more precise segmentation of tiny anomalies.

\section{Discussion}
\label{sec:discussion}

\begin{table}[!t]
    \centering
  \setlength{\tabcolsep}{0.8mm}{
    \begin{tabular}{cc|c}
    \toprule
    Swin-Cropping & Data augmentation & F1-max \\
    \midrule
    \checkmark &  & 33.6 \\
     & \checkmark & 31.8 \\
    \checkmark & \checkmark & \textbf{35.5} \\
    \bottomrule
    \end{tabular}
    }
\caption{Ablation of our Swin-Cropping and data augmentation operation.
We report the F1-score with optimal threshold (F1-max in \%) to measure the performance of anomaly maps on $\textit{TEST}_\text{pub}$.}
\label{abl:amap}
\end{table}

\begin{table}[!t]
    \centering
  \setlength{\tabcolsep}{14pt}{
    \begin{tabular}{cc|c}
    \toprule
    MEBin & SAM-Finer & SegF1 \\
    \midrule
    \checkmark &  & 34.4 \\
     & \checkmark & 46.5 \\
    \checkmark & \checkmark & \textbf{46.9} \\
    \bottomrule
    \end{tabular}
    }
\caption{Ablation of MEBin and SAM-Finer.
We report the SegF1 (in \%) to measure the performance of binary masks on $\textit{TEST}_\text{pub}$.}
\label{abl:mask}
\end{table}

\subsection{Ablation study}
Table \ref{abl:amap} presents ablation studies of our Swin-Cropping and data augmentation modules on the $\textit{TEST}_\text{pub}$ set.
Due to the direct impact on anomaly map quality, we evaluate performance using the F1-max metric, which indicates the theoretical upper bound of SegF1 under the optimal threshold.
Our analysis reveals that Swin-Cropping achieves 3.7\% F1-max improvement by preserving critical anomaly information in sub-images, 
effectively mitigating missing detections of small anomalies.
The data augmentation module contributes an additional 1.9\% gain through exposure variation simulation and Gaussian noise injection,
demonstrating enhanced robustness against various lighting conditions.

In Table \ref{abl:mask}, we conduct the ablation studies of the MEBin and SAM-Finer modules, which directly affect the binarization performance of anomaly maps.
The MEBin module achieves a 0.4\% SegF1 improvement by adaptively detecting subtle anomalies overlooked by conventional "mean+3std" thresholding, reducing false negatives.
SAM-Finer delivers a substantial 12.5\% performance gain, primarily through internal defect filling in fabric-class anomalies where many false negatives exist.
While SAM enables fine-grained refinement of slender and small anomalies, its impact on quantitative metrics remains limited.

\subsection{Future work}
While our framework demonstrates state-of-the-art performance, several promising directions emerge for advancing industrial anomaly detection in real-world scenarios.
Firstly, the detection of small anomalies in high-resolution industrial images remains excessive time consumption.
Studying an efficient AD model for processing high-resolution images is necessary.
Meanwhile, the adaptive binarization is a direction worth exploring.
However, there are relatively few studies on this at present.
It is promising to explore a binarization strategy that is robust to different product categories.

\section{Conclusion}
In this paper, we present our solution RoBiS for the CVPR VAND3.0 challenge in the MVTec AD 2 dataset.
Three important modules are contained in our RoBiS:
(1) Swin-Cropping to pre-processing high-resolution images.
(2) Data augmentation to simulate different lighting conditions and then training the AD model INP-Former.
(3) Adaptive binarization and SAM-Finer to generate high-quality binary masks.
Our solution achieves 51.00\% and 46.52\% SegF1 in the $\textit{TEST}_\text{priv}$ and $\textit{TEST}_\text{priv,mix}$, respectively.

\section{Acknowledgments}
\noindent This work was supported by the National Natural Science Foundation of China under Grant No.62176098. The computation is completed in the HPC Platform of Huazhong University of Science and Technology.

\newpage

{
    \small
    \bibliographystyle{ieeenat_fullname}
    \bibliography{main}
}

\end{document}